\documentclass{article}

\usepackage[final,nonatbib]{neurips_2018}

\usepackage[natbib=true,backend=biber,giveninits=true,maxbibnames=10,maxcitenames=2,sortcites=true]{biblatex}
\bibliography{references.bib}

\usepackage[utf8]{inputenc}
\usepackage[T1]{fontenc}
\usepackage{hyperref}
\usepackage{url}
\usepackage{booktabs}
\usepackage{amsfonts}
\usepackage{nicefrac}
\usepackage{microtype}

\usepackage{amsmath}
\usepackage{amssymb}
\usepackage{amsthm}
\usepackage{bm}
\usepackage{subcaption}
\usepackage{graphicx}
\usepackage{multirow}
\usepackage{xcolor}
\usepackage{enumitem}

\usepackage{siunitx}
\DeclareMathOperator{\E}{\mathbb{E}}
\let\vec\bm

\newcommand{\wrt}{w.r.t.~}
\newcommand{\eg}{e.g.~}
\newcommand{\Sz}{\mathcal{S}_{\vec{\phi}}(\vec{z})}
\newcommand{\Szscalar}{\mathcal{S}_{\vec{\phi}}(z)}
\newcommand{\invS}{\mathcal{S}^{-1}_{\vec{\phi}}(\vec{\varepsilon})}
\newcommand{\SNormal}{\mathcal{S}_{\mu, \sigma}(z)}
\newcommand{\invSNormal}{\mathcal{S}^{-1}_{\mu, \sigma}(\varepsilon)}

\title{Implicit Reparameterization Gradients}
\author{
  Michael Figurnov \quad Shakir Mohamed \quad Andriy Mnih\\
  DeepMind, London, UK \\
  \texttt{\{mfigurnov,shakir,amnih\}@google.com}
}

\begin{document}

\maketitle

\begin{abstract}
By providing a simple and efficient way of computing low-variance gradients of continuous random variables, the reparameterization trick has become the technique of choice for training a variety of latent variable models. However, it is not applicable to a number of important continuous distributions.  We introduce an alternative approach to computing reparameterization gradients based on implicit differentiation and demonstrate its broader applicability by applying it to Gamma, Beta, Dirichlet, and von Mises distributions, which cannot be used with the classic reparameterization trick. Our experiments show that the proposed approach is faster and more accurate than the existing gradient estimators for these distributions.
\end{abstract}

\section{Introduction}
Pathwise gradient estimators are a core tool for stochastic estimation in machine learning and statistics \cite{fu2006gradient, glasserman2013monte, titsias2014doubly, kingma2014auto, rezende2014stochastic}. In machine learning, we now commonly introduce these estimators using the ``reparameterization trick'', in which we replace a probability distribution with an equivalent parameterization of it, using a deterministic and differentiable transformation of some fixed base distribution. This reparameterization is a powerful tool for learning because it makes backpropagation possible in computation graphs with certain types of continuous random variables, e.g. with Normal, Logistic, or Concrete distributions \cite{jang2017categorical, maddison2017concrete}. Many of the recent advances in machine learning were made possible by this ability to backpropagate through stochastic nodes. They include variational autoenecoders (VAEs), automatic variational inference \cite{kingma2014auto, rezende2014stochastic, kucukelbir2017automatic}, Bayesian learning in neural networks \cite{blundell2015weight, gal2016dropout}, and principled regularization in deep networks \cite{gal2016theoretically, molchanov2017variational}.

The reparameterization trick is easily used with distributions that have location-scale parameterizations or tractable inverse cumulative distribution functions (CDFs), or are expressible as deterministic transformations of such distributions.
These seemingly modest requirements are still fairly restrictive as they preclude a number of standard distributions, such as truncated, mixture, Gamma, Beta, Dirichlet, or von Mises, from being used with reparameterization gradients. This paper provides a general tool for reparameterization in these important cases.

The limited applicability of reparameterization has often been addressed by using a different class of gradient estimators, the score-function estimators \citep{fu2006gradient, glynn1990likelihood, williams1992simple}. While being more general, they typically result in high-variance gradients which require problem-specific variance reduction techniques to be practical.
Generalized reparameterizations involve combining the reparameterization and score-function estimators \citep{ruiz2016generalized, naesseth2017reparameterization}.
Another approach is to approximate the intractable derivative of the inverse CDF \cite{knowles2015stochastic}.

Following~\citet{graves2016stochastic}, we use implicit differentiation to differentiate the CDF rather than its inverse.
While the method of \citet{graves2016stochastic} is only practical for distributions with analytically tractable CDFs and has been used solely with mixture distributions, we leverage automatic differentiation to handle distributions with \emph{numerically} tractable CDFs, such as Gamma and von Mises.
We review the standard reparameterization trick in Section \ref{sect:standard_trick} and then make the following contributions:
\vspace{-2mm}
\begin{itemize}[leftmargin=*,noitemsep]
\item We develop \emph{implicit reparameterization gradients} that provide unbiased estimators for continuous distributions with numerically tractable CDFs. This allows many other important distributions to be used as easily as the Normal distribution in stochastic computation graphs.
\item We show that the proposed gradients are both faster and more accurate than alternative approaches.
\item We demonstrate that our method can outperform existing stochastic variational methods at training the Latent Dirichlet Allocation topic model in a black-box fashion using amortized inference.
\item We use implicit reparameterization gradients to train VAEs with Gamma, Beta, and von Mises latent variables instead of the usual Normal variables, leading to latent spaces with interesting alternative topologies.
\end{itemize}

\section{Background}
\label{sect:standard_trick}

\subsection{Explicit reparameterization gradients}
\label{sec:reparam}
We start with a review of the original formulation of reparameterization gradients~\cite{titsias2014doubly,kingma2014auto,rezende2014stochastic}, which we will refer to as  \emph{explicit} reparameterization.
Suppose we would like to optimize an expectation $\E_{q_{\vec{\phi}} (\vec{z})}\left[f(\vec{z})\right]$ of some continuously differentiable function $f(\vec{z})$ \wrt the parameters $\vec{\phi}$ of the distribution.
We assume that we can find a standardization function $\Sz$ that when applied to a sample from $q_{\vec{\phi}} (\vec{z})$ removes its dependence on the parameters of the distribution. The standardization function should be continuously differentiable \wrt its argument and parameters, and invertible:
\begin{equation}
    \Sz = \vec{\varepsilon} \sim q(\vec{\varepsilon}) \qquad \vec{z} = \invS. \label{eqn:z-g-inverse}
\end{equation}

For example, for a Gaussian distribution $\mathcal{N}(\mu, \sigma)$ we can use $\SNormal = (z - \mu)/\sigma \sim \mathcal{N}(0, 1)$.
We can then express the objective as an expectation \wrt $\vec{\varepsilon}$, transferring the dependence on $\vec{\phi}$ into $f$:
\begin{equation}
\E_{q_{\vec{\phi}} (\vec{z})} \left[f(\vec{z})\right] = \E_{q(\vec{\varepsilon})} \left[ f(\invS) \right].
\end{equation}
This allows us to compute the gradient of the expectation as the expectation of the gradients:
\begin{equation}
    \nabla_{\vec{\phi}} \E_{q_{\vec{\phi}} (\vec{z})} \left[f(\vec{z})\right] = \E_{q(\vec{\varepsilon})} \left[\nabla_{\vec{\phi}} f(\invS) \right] = \E_{q(\vec{\varepsilon})} \left[ \nabla_{\vec{z}} f(\invS) \nabla_{\vec{\phi}} \invS \right].
    \label{eqn:explicit-reparameterization}
\end{equation}
A standardization function $\Sz$ satisfying the requirements exists for a wide range of continuous distributions, but it is not always practical to take advantage of this.
For instance, the CDF $F(z | \vec{\phi})$ of a univariate distribution provides such a function, mapping samples from it to samples from the uniform distribution over $[0, 1]$.
However, inverting the CDF is often complicated and expensive, and computing its derivative is even harder.

\subsection{Stochastic variational inference}

Stochastic variational inference~\cite{hoffman2013stochastic} for latent variable models is perhaps the most popular use case for reparameterization gradients. Consider a model $p_{\vec{\theta}} (\vec{x}) = \int p_{\vec{\theta}} (\vec{x} | \vec{z}) p(\vec{z}) d\vec{z}$, where $\vec{x}$ is an observation, $\vec{z} \in \mathbb{R}^D$ is a vector-valued latent variable, $p_{\vec{\theta}} (\vec{x}|\vec{z})$ is the likelihood function with parameters $\vec{\theta}$, and $p(\vec{z})$ is the prior distribution. Except for a few special cases, maximum likelihood learning in such models is intractable because of the difficulty of the integrals involved. Variational inference~\cite{jaakkola2000bayesian} provides a tractable alternative by introducing a variational posterior distribution $q_{\vec{\phi}} (\vec{z} | \vec{x})$ and maximizing a lower bound on the marginal log-likelihood:
\begin{equation}
    \mathcal{L}(\vec{x}, \vec{\theta}, \vec{\phi}) = \E_{q_{\vec{\phi}} (\vec{z} | \vec{x})} \left[ \log p_{\vec{\theta}} (\vec{x} | \vec{z})\right] - \operatorname{KL}(q_{\vec{\phi}} (\vec{z} | \vec{x}) \| p(\vec{z}) ) \leq \log p_{\vec{\theta}} (\vec{x}).
\end{equation}
Training models with modern stochastic variational inference \cite{paisley2012variational,kingma2014auto} involves gradient-based optimization of the bound \wrt the model parameters $\vec{\theta}$ and the variational posterior parameters $\vec{\phi}$. While the KL-divergence term and its gradients can often be computed analytically, the remaining term and its gradients are typically intractable and are approximated using samples from the variational posterior. The most general form of this approach involves score-function gradient estimators \cite{paisley2012variational,ranganath2014black,mnih2014neural} that handle both discrete and continuous latent variables but have relatively high variance. The reparameterization trick usually provides a lower variance gradient estimator and is easier to use, but due to the limitations discussed above, is not applicable to many important continuous distributions.

\section{Implicit reparameterization gradients}

We propose an alternative way of computing the reparameterization gradient that avoids the inversion of the standardization function.
We start from Eqn.~\eqref{eqn:explicit-reparameterization} and perform a change of variable $\vec{z} = \invS$:
\begin{equation}
    \nabla_{\vec{\phi}} \E_{q_{\vec{\phi}} (\vec{z})} \left[f(\vec{z})\right] = \E_{q_{\vec{\phi}} (\vec{z})} \left[\nabla_{\vec{z}} f(\vec{z}) \nabla_{\vec{\phi}} \vec{z}\right]; \qquad \nabla_{\vec{\phi}} \vec{z} = \nabla_{\vec{\phi}} \invS |_{\vec{\varepsilon} = \Sz}.
    \label{eqn:explicit-grad-z}
\end{equation}
Our key insight is that we can compute $\nabla_{\vec{\phi}} \vec{z}$ by \emph{implicit differentiation}.
We apply the \emph{total} gradient $\nabla^{\mathrm{TD}}_{\vec{\phi}}$ to the equality $\Sz = \vec{\varepsilon}$.
Then, we use the chain rule to expand the total gradient in terms of the partial gradients.
The standardization function $\Sz$ depends on the parameters $\vec{\phi}$ directly via the subscript parameters and indirectly via the argument $\vec{z}$, while the noise $\vec{\varepsilon}$ is independent of $\vec{\phi}$ by the definition of a standardization function.
Thus, we have
$\nabla_{\vec{z}} \Sz \nabla_{\vec{\phi}} \vec{z} + \nabla_{\vec{\phi}} \Sz = \vec{0}$,
where all the gradients are partial.
Solving this equation for $\nabla_{\vec{\phi}} \vec{z}$ yields
\begin{equation}
    \boxed{ \nabla_{\vec{\phi}} \vec{z} = -(\nabla_{\vec{z}} \Sz)^{-1} \nabla_{\vec{\phi}} \Sz }
    \label{eqn:z-grad}
\end{equation}
This expression for the gradient only requires differentiating the standardization function and not inverting it.
Note that its value does not change under any invertible transformation $T(\vec{\varepsilon})$ of the standardization function, since the corresponding Jacobian $\nabla_{\vec{\varepsilon}} T(\vec{\varepsilon})$ cancels out with the inverse.

\textbf{Example: univariate Normal distribution} $\mathcal{N}(\mu, \sigma^2)$.
We illustrate that explicit and implicit reparameterizations give identical results.
A standardization function is given by $\SNormal = (z - \mu)/\sigma = \varepsilon \sim \mathcal{N}(0, 1)$.
Explicit reparameterization inverts this function: $z = \invSNormal = \mu + \sigma \varepsilon, \ \frac{\partial z}{\partial \mu} = 1, \ \frac{\partial z}{\partial \sigma} = \varepsilon$.
The implicit reparameterization, Eqn.~\eqref{eqn:z-grad}, gives:
\begin{equation}
    \frac{\partial z}{\partial \mu} = -\frac{ \frac{\partial \SNormal} {\partial \mu} }{\frac{\partial \SNormal}{\partial z}} = -\frac{-\frac{1}{\sigma}}{\frac{1}{\sigma}} = 1, \quad \frac{\partial z}{\partial \sigma} = -\frac{ \frac{\partial \SNormal} {\partial \sigma} }{\frac{\partial \SNormal}{\partial z}} = -\frac{-\frac{z - \mu}{\sigma^2}}{\frac{1}{\sigma}} = \frac{z - \mu}{\sigma}.
\end{equation}
The expressions are equivalent, but the implicit version avoids inverting $\SNormal$.

\textbf{Universal standardization function.}
For univariate distributions, a standardization function is given by the CDF: $\Szscalar = F(z | \vec{\phi}) \sim \operatorname{Uniform}(0, 1)$.
Assuming that the CDF is strictly monotonic and continuously differentiable \wrt $z$ and $\vec{\phi$}, it satisfies the requirements for a standardization function.
Plugging this function into~\eqref{eqn:z-grad}, we have
\begin{equation}
    \nabla_{\vec{\phi}} z = - \frac{\nabla_{\vec{\phi}} F(z | \vec{\phi})}{q_{\vec{\phi}} (z)}.
    \label{eqn:z-grad-univariate}
\end{equation}
Therefore, computing the implicit gradient requires only differentiating the CDF.
In the multivariate case, we can perform the multivariate distributional transform~\cite{ruschendorf2013copulas}:
\begin{equation}
    \Sz = ( F(z_1 | \vec{\phi}), F(z_2 | z_1, \vec{\phi}), \dots, F(z_D | z_1, \dots, z_{D-1}, \vec{\phi}) ) = \vec{\varepsilon},
\end{equation}
where $q(\vec{\varepsilon}) = \prod_{d=1}^D \operatorname{Uniform}(\varepsilon_d | 0, 1)$.
Eqn.~\eqref{eqn:z-grad} requires computing the gradient of the (conditional) CDFs and solving a linear system with matrix $\nabla_{\vec{z}} \Sz$.
If the distribution is factorized, the matrix is diagonal and the system can be solved in $O(D)$.
Otherwise, the matrix is triangular because each CDF depends only on the preceding elements, and the system is solvable in $O(D^2)$.

\textbf{Algorithm.}
We present the comparison between the standard explicit and the proposed implicit reparameterization in Table~\ref{fig:reparameterization-algos}.
Samples of $\vec{z}$ in implicit reparameterization can be obtained with any suitable method, such as rejection sampling~\cite{devroye1986}.
The required gradients of the standardization function can be computed either analytically or using automatic differentiation.

\begin{table}[ht]
    \centering
    \caption{Comparison of the two reparameterization types. While they provide the same result, the implicit version is easier to implement for distributions such as Gamma because it does not require inverting the standardization function $\Sz$.}
    \begin{tabular}{lll}\toprule
        & Explicit reparameterization & Implicit reparameterization (proposed) \\ \midrule
        \multirow{2}{*}{Forward pass} & Sample $\vec{\varepsilon} \sim q(\vec{\varepsilon})$ & Sample $\vec{z} \sim q_{\vec{\phi}}(\vec{z})$ \\
        & Set $\vec{z} \gets \invS$ & \\ \midrule
        \multirow{2}{*}{Backward pass}  & Set $\nabla_{\vec{\phi}} \vec{z} \gets \nabla_{\vec{\phi}} \invS$ & Set $\nabla_{\vec{\phi}} \vec{z} \gets -(\nabla_{\vec{z}} \Sz)^{-1} \nabla_{\vec{\phi}} \Sz$ \\
        & Set $\nabla_{\vec{\phi}} f(\vec{z}) \gets \nabla_{\vec{z}} f(\vec{z}) \nabla_{\vec{\phi}} \vec{z}$ & Set $\nabla_{\vec{\phi}} f(\vec{z}) \gets \nabla_{\vec{z}} f(\vec{z})  \nabla_{\vec{\phi}} \vec{z}$ \\
     \bottomrule
    \end{tabular}
    \label{fig:reparameterization-algos}
\end{table}

\section{Applications of implicit reparameterization gradients}

We now demonstrate how implicit reparameterization can be applied to a variety of distributions.
Our strategy is to provide a computation method for a standardization function, such as CDF or multivariate distributional transform, and its gradients.

\textbf{Truncated univariate distribution}.
A truncated distribution is obtained by restricting a distribution's domain to some range $[a, b]$.
Its CDF can be computed from the CDF of the original distribution: $\hat{F}(z | \vec{\phi}, a, b) = \frac{F(z | \vec{\phi}) - F(a | \vec{\phi})}{F(b | \vec{\phi}) - F(a | \vec{\phi})}, \ z \in [a, b]$.
Assuming that the gradient $\nabla_{\vec{\phi}} F(z | \vec{\phi})$ is available, we can easily compute the implicit gradient for the truncated distribution.

\textbf{Mixture distribution} $q_{\vec{\phi}} (\vec{z}) = \sum_{i=1}^K w_i q_{\vec{\phi}_i} (\vec{z})$, where $\vec{\phi} = (\vec{\phi}_1, \dots, \vec{\phi}_K, w_1, \dots, w_K)$.
In the univariate case, the CDF of the mixture is simply $\sum_{i=1}^K w_i F(z | \vec{\phi}_i)$.
In the multivariate case, the distributional transform is given by $F(z_d | z_1, \dots, z_{d-1}, \vec{\phi}) = \sum_{i=1}^K w^d_i F(z_d | z_1, \dots, z_{d-1}, \vec{\phi_i})$,
where $w^d_i = \frac{w_i q_{\vec{\phi}_i}(z_1, \dots, z_{d-1})}{\sum_{j=1}^K w_j q_{\vec{\phi}_j}(z_1, \dots, z_{d-1})}$ is the posterior weight for the mixture component after observing the first $d-1$ dimensions of the sample.
The required gradient can be obtained via automatic differentiation.
When the mixture components are fully factorized, we obtain the same result as~\cite{graves2016stochastic}, but in a simpler form, due to automatic differentiation and the explicitly specified linear system.

\textbf{Gamma distribution} $\operatorname{Gamma}(\alpha, \beta)$ with shape $\alpha > 0$ and rate $\beta > 0$.
The rate can be standardized using the scaling property: if $z \sim \operatorname{Gamma}(\alpha, 1)$, then $z/\beta \sim \operatorname{Gamma}(\alpha, \beta)$.
For the shape parameter, the CDF of the Gamma distribution with shape $\alpha$ and unit rate is the regularized incomplete Gamma function $\gamma(z, \alpha)$ that does not have an analytic expression.
Following~\citet{moore1982derincgamma}, we propose to apply forward-mode automatic differentiation~\cite{baydin2015automatic} to a numerical method~\cite{bhattacharjee1970incgamma} that computes its value.
This provides the derivative $\frac{\partial \gamma(z, \alpha)}{\partial \alpha}$ for roughly twice the cost of computing the CDF.

\textbf{Student's $t$-distribution} samples can be derived from samples of Gamma.
Indeed, if $\sigma \sim \operatorname{Gamma}(\frac{\nu}{2}, \frac{\nu}{2})$, then $z \sim \mathcal{N}(0, \sigma^2)$ is $t$-distributed with $\nu$ degrees of freedom.

\textbf{Beta and Dirichlet distribution} samples can also be obtained from samples of Gamma.
If $z_1 \sim \operatorname{Gamma}(\alpha, 1)$ and $z_2 \sim \operatorname{Gamma}(\beta, 1)$, then $\frac{z_1}{z_1 + z_2} \sim \operatorname{Beta}(\alpha, \beta)$.
Similarly, if $z_i \sim \operatorname{Gamma}(\alpha_i, 1)$, then $\left(\frac{z_1}{\sum_{j=1}^D z_j}, \dots, \frac{z_D}{\sum_{j=1}^D z_j}\right) \sim \operatorname{Dirichlet}(\alpha_1, \dots, \alpha_D)$.

\textbf{Von Mises distribution}~\cite{von1918uber,mardia2009directional} is a maximum entropy distribution on a circle with the density function $\operatorname{vonMises}(z | \mu, \kappa) = \frac{\exp(\kappa \cos(z - \mu))}{2 \pi I_0 (\kappa)}$, where $\mu$ is the location parameter, $\kappa > 0$ is the concentration, and $I_0(\kappa)$ is the modified Bessel function of the first kind.
The location parameter $\mu$ can be standardized by noting that if $z \sim \operatorname{vonMises}(0, \kappa)$, then $z + \mu \sim \operatorname{vonMises}(\mu, \kappa)$.
For the concentration parameter $\kappa$, we propose to use implicit reparameterization by performing forward-mode automatic differentiation of an efficient numerical method~\cite{hill1977algorithm} for computation of the CDF.

\subsection{Accuracy and speed of reparameterization gradient estimators}
Implicit reparameterization requires differentiating the CDF \wrt its parameters.
When this operation is analytically intractable, \eg for Gamma and von Mises distributions, we estimate it via forward-mode  differentiation of the code that numerically evaluates the CDF.
We implement this approach by manually performing the required modifications of the C++ code (see Appendix~\ref{sec:reparameterization-gradient-implementation}).
An alternative is to use a central finite difference approximation of the derivative: $\frac{\partial F(z | \phi)}{\partial \phi} \approx \frac{F(z | \phi (1 + \delta)) - F(z | \phi (1 - \delta))}{2 \phi \delta}$, where $0 < \delta < 1$ is the \emph{relative} step size that we choose via grid search.
For the Gamma distribution, we also compare with two alternatives: (1) the estimator of~\citet{knowles2015stochastic} that performs explicit reparameterization by approximately computing the derivative of the inverse CDF; (2) the concurrently developed method of~\citet{jankowiak2018pathwisebeyond} that computes implicit reparameterization using a closed-form approximation of the CDF derivative.
We use the reference PyTorch~\citet{paszke2017automatic} implementation of the method of~\citet{jankowiak2018pathwisebeyond}.
The ground truth value of the CDF derivative is computed in a computationally expensive but accurate way (see Appendix~\ref{sec:accuracy-speed-appendix}).
The results in Table~\ref{fig:gradient-accuracy} suggest that the automatic differentiation approach provides the highest accuracy and speed.
The finite difference method can be easier to implement if a CDF computation method is available, but requires computation in \texttt{float64} to obtain the \texttt{float32} precision.
This can be problematic for devices such as GPUs and other accelerators that do not support fast high-precision computation.
The approach of Knowles is slower and significantly less accurate due to the approximations of the inverse CDF derivative computation method.
The method of Jankowiak and Obermeyer is $4.5\times$ slower and $3\times$ less accurate than the automatic differentiation approach, which reflects the complexity of obtaining fast and accurate closed-form approximations to the CDF derivative.
In the remaining experiments we use automatic differentiation and \texttt{float32} precision.

\begin{table}
    \centering
    \caption{Average error and time (measured in seconds per element) of the reparameterization gradient computation methods. Automatic differentiation achieves the lowest error and the highest speed.}
    \resizebox{\linewidth}{!}{
    \begin{tabular}{llcccc}\toprule
        & & \multicolumn{2}{c}{\textbf{Gamma}} & \multicolumn{2}{c}{\textbf{Von Mises}} \\
        Method & Precision & Mean abs. error & Time (s) & Mean abs. error & Time (s) \\ \midrule
        Automatic differentiation & \multirow{3}{*}{\texttt{float32}} & \textbf{\num{2.3e-06}} & \textbf{\num{1.9e-08}} & \textbf{\num{1.9e-07}} & \textbf{\num{3.1e-08}} \\
        Finite difference & & \num{1.9e-03} & \num{3.8e-08} & \num{9.6e-05} & \num{3.8e-08} \\
        \citet{jankowiak2018pathwisebeyond} & & \num{4.1e-05} & \num{9.0e-08} & -- & -- \\
        \midrule
        Automatic differentiation & \multirow{3}{*}{\texttt{float64}} & \textbf{\num{5.4e-13}} & \textbf{\num{3.2e-08}} & \textbf{\num{1.3e-13}} & \textbf{\num{3.7e-08}} \\
        Finite difference & & \num{3.2e-09} & \num{7.1e-08} & \num{1.1e-10} & \num{5.9e-08} \\
        \citet{knowles2015stochastic} & & \num{6.5e-03} & \num{1.2e-06} & -- & -- \\
     \bottomrule
    \end{tabular}
    }
    \label{fig:gradient-accuracy}
\end{table} 

\section{Related work}
\label{sect:related}

\textbf{Surrogate distributions.}
When explicit reparameterization is not feasible, it is often possible to modify the model to use alternative distributions that are reparameterizable. This is a popular approach due to is simplicity.
\citet{kucukelbir2017automatic} approximate posterior distributions by a deterministic transformation of Normal samples; \citet{nalisnick2016approximate,nalisnick2017stick} replace Beta distributions with Kumaraswamy distributions in the Dirichlet Process stick-breaking construction; \citet{zhang2018whai} substitute the Gamma distribution for a Weibull distribution; \citet{srivastava2017autoencoding,srivastava2018variational} replace the Dirichlet distribution with a Logistic Normal.
Surrogate distributions however do not always have all the desirable properties of the distributions they replace. For example, as noted by~\citet{ruiz2016generalized}, such surrogate distributions struggle to capture sparsity, which is achievable with Gamma and Dirichlet distributions.

\textbf{Integrating out the nuisance variables.}
In some cases it is possible to trade computation for simplicity of reparameterization.
\citet{roeder2017sticking} consider a mixture of reparameterizable distributions and analytically sum out the discrete mixture component id variable.
For a mixture with $K$ components, this results in a $K$-fold increase of computation, compared to direct reparameterization of the mixture.
This approach becomes prohibitively expensive for a chain of mixture distributions, where the amount of computation grows exponentially with the length of the chain.
On the other hand, we can always estimate the gradients with just one sample by directly reparameterizing the mixture.

\textbf{Implicit reparameterization gradients.}
Reparameterization gradients have been known in the operations research community since the late 1980s under the name of pathwise, or stochastic, gradients~\citep{suri1988perturbation,fu2006gradient}.
There the ``explicit'' and ``implicit'' versions were usually introduced side-by-side, but they were applied only to univariate distributions and simple computational graphs that do not require backpropagation.
In the machine learning community, the implicit reparameterization gradients for univariate distributions were introduced by~\citet{salimans2013fixed}.
That work, as well as \citet{hoffman2015stochastic}, used the implicit gradients to perform backpropagation through the Gamma distribution using a finite difference approximation of the CDF derivative.
\citet{graves2016stochastic} independently introduced the implicit reparameterization gradients for multivariate distributions with analytically tractable CDFs, such as mixtures. We add to this rich literature by generalizing the technique to handle arbitrary standardization functions, deriving a simpler expression than that of \citet{graves2016stochastic} for the multivariate case, showing the connection to explicit reparameterization gradients, and providing an efficient automatic differentiation method to compute the intractable CDF derivatives.

\textbf{Reparameterization gradients as differential equation solutions.}
The concurrent works \citep{jankowiak2018pathwisebeyond,jankowiak2018pathwisemultivariate} provide a complementary view of the reparameterization gradients as solutions of a differential equation called the transport equation.
For univariate distributions, the unique solution is Eqn.~\eqref{eqn:z-grad-univariate}.
However, for the non-factorial multivariate distributions, there are multiple solutions.
By choosing an appropriate one, the variance of the gradient estimator may be reduced.
Unfortunately, there does not seem to be a general way to obtain these solutions, so distribution-specific derivations are required.
We hypothesize that the transport equation solutions correspond to the implicit reparameterization gradients for different standardization functions.

\textbf{Generalized reparameterizations.}
The limitations of standard reparameterization was recently tackled by several other works. 
\citet{ruiz2016generalized} introduced generalized reparameterization gradients (GRG) that expand the applicability of the reparameterization trick by using a standardization function that allows the underlying base distribution to depend weakly on the parameter vector (\eg only through the higher moments). The resulting gradient estimator, which in addition to the the reparameterized gradients term includes a score-function gradient term that takes into account the dependence of the base distribution on the parameter vector, was applied to the Gamma, Beta, and log-Normal distributions. The challenge of using this approach lies in finding an effective approximate standardization function, which is nontrivial yet essential for obtaining low-variance gradients.

Rejection sampling variational inference (RSVI) \citep{naesseth2017reparameterization} is a closely-related approach that combines the reparameterization gradients from the proposal distribution of a rejection sampler with a score-function gradient term that takes into account the effect of the accept/reject step. When applied to the gamma distribution the RSVI gradients can have lower variance gradients than those computed using GRG \citep{naesseth2017reparameterization}.
\citet{davidson2018hyperspherical} have recently demonstrated the use of RSVI with the von Mises-Fisher distribution.

\section{Experiments}

We apply implicit reparameterization for two distributions with analytically intractable CDFs (Gamma and von Mises) to three problems: a toy setting of stochastic cross-entropy estimation, training a Latent Dirichlet Allocation~\cite{blei2003latent} (LDA) topic model, and training VAEs~\cite{kingma2014auto,rezende2014stochastic} with non-Normal latent distributions.
We use the RSVI gradient estimator~\cite{naesseth2017reparameterization} as our main baseline.
For Gamma distributions, RSVI provides a \emph{shape augmentation} parameter $B$ that decreases the magnitude of the score-function correction term by using additional $B$ samples from a uniform distribution.
As $B \to \infty$, the term vanishes and the RSVI gradient becomes equivalent to ours, but with a higher computational cost.
Von Mises distribution does not have such an augmentation parameter.
For LDA, we also compare with a surrogate distribution approach~\cite{srivastava2017autoencoding} and a classic stochastic variational inference method~\cite{hoffman2013stochastic}.
The experimental details are given in Appendix~\ref{sec:experimental-details}.
We use TensorFlow~\cite{abadi2016tensorflow} for our experiments.
Implicit reparameterization for Gamma, Student's $t$, Beta, Dirichlet and von Mises distributions is available in TensorFlow Probability~\cite{dillon2017tensorflow}.
This library also contains an implementation of the LDA model from section~\ref{sec:lda}.

\subsection{Gradient of the cross-entropy}

We compare the variance of the implicit and RSVI gradient estimators on a toy problem of stochastic estimation of the cross-entropy gradient, $\frac{\partial}{\partial \phi} \E_{q_{\phi} (\vec{z})} [-\log p(\vec{z})]$.
It was introduced by~\citet{naesseth2017reparameterization} as minimization of the KL-divergence; however, since they analytically compute the entropy, the only source of variance is the cross-entropy term.
We use their setting for the Dirichlet distribution: $p(\vec{z}) = \operatorname{Dirichlet}(\vec{z} | \alpha_1, \alpha_2, \dots, \alpha_{100}),\ q_{\phi} (\vec{z}) = \operatorname{Dirichlet}(\vec{z} | \phi, \alpha_2, \dots, \alpha_{100})$, where $\vec{\alpha}$ are the posterior parameters for a Dirichlet with a uniform prior after observing $100$ samples from a Categorical distribution.
The Dirichlet samples are obtained by transforming samples from Gamma.
Additionally, we construct a similar problem with the von Mises distribution: $p(\vec{z}) = \prod_{d=1}^{10} \operatorname{vonMises}(z_d | 0, 2)$ and $q_{\phi} (\vec{z}) = \operatorname{vonMises}(z_1 | 0, \phi) \prod_{d=2}^{10} \operatorname{vonMises}(z_d | 0, 2)$.

\begin{figure}
    \centering
    \begin{subfigure}[b]{0.31\linewidth}
        \includegraphics[width=\linewidth]{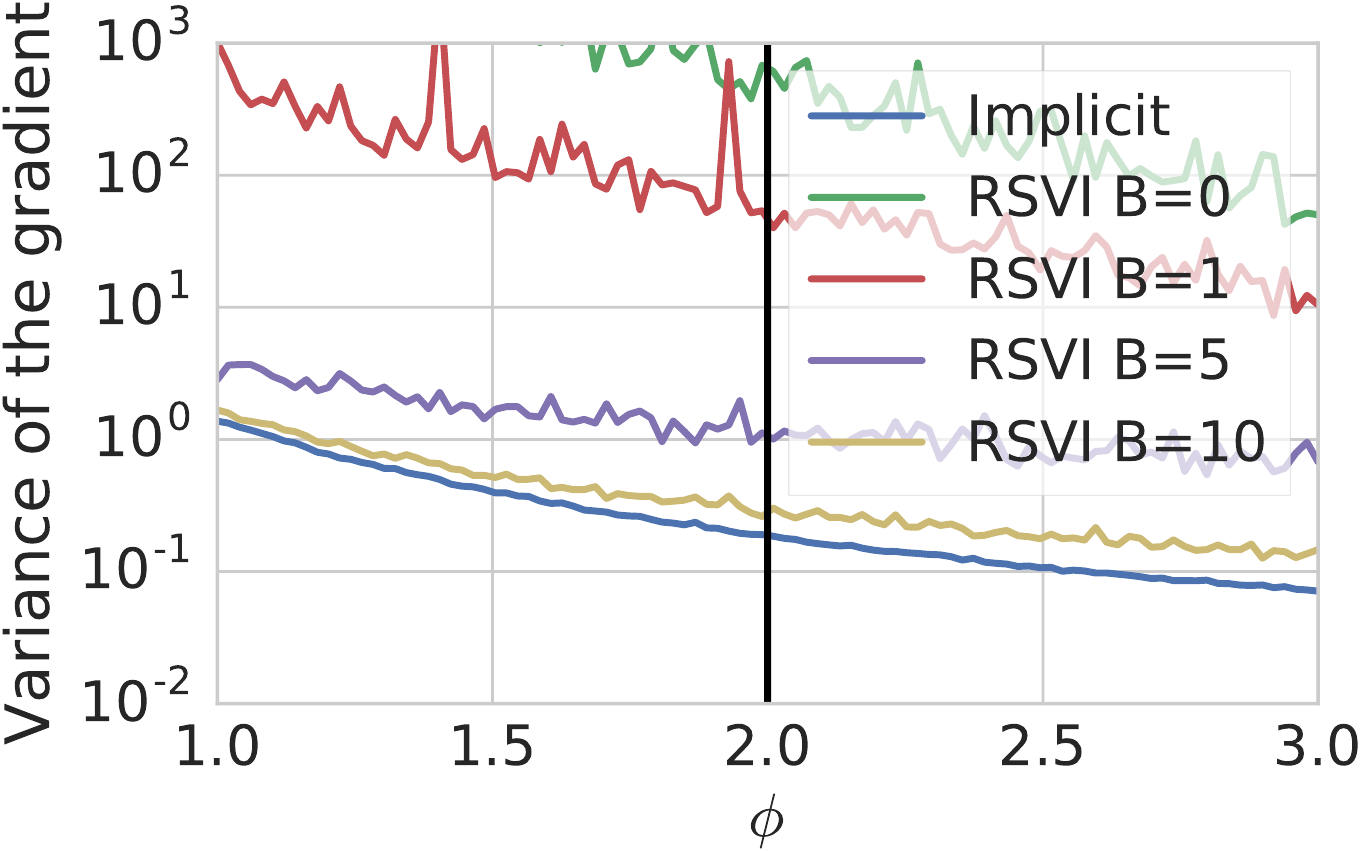}
        \caption{Dirichlet distribution}
    \end{subfigure}
    \begin{subfigure}[b]{0.31\linewidth}
        \includegraphics[width=\linewidth]{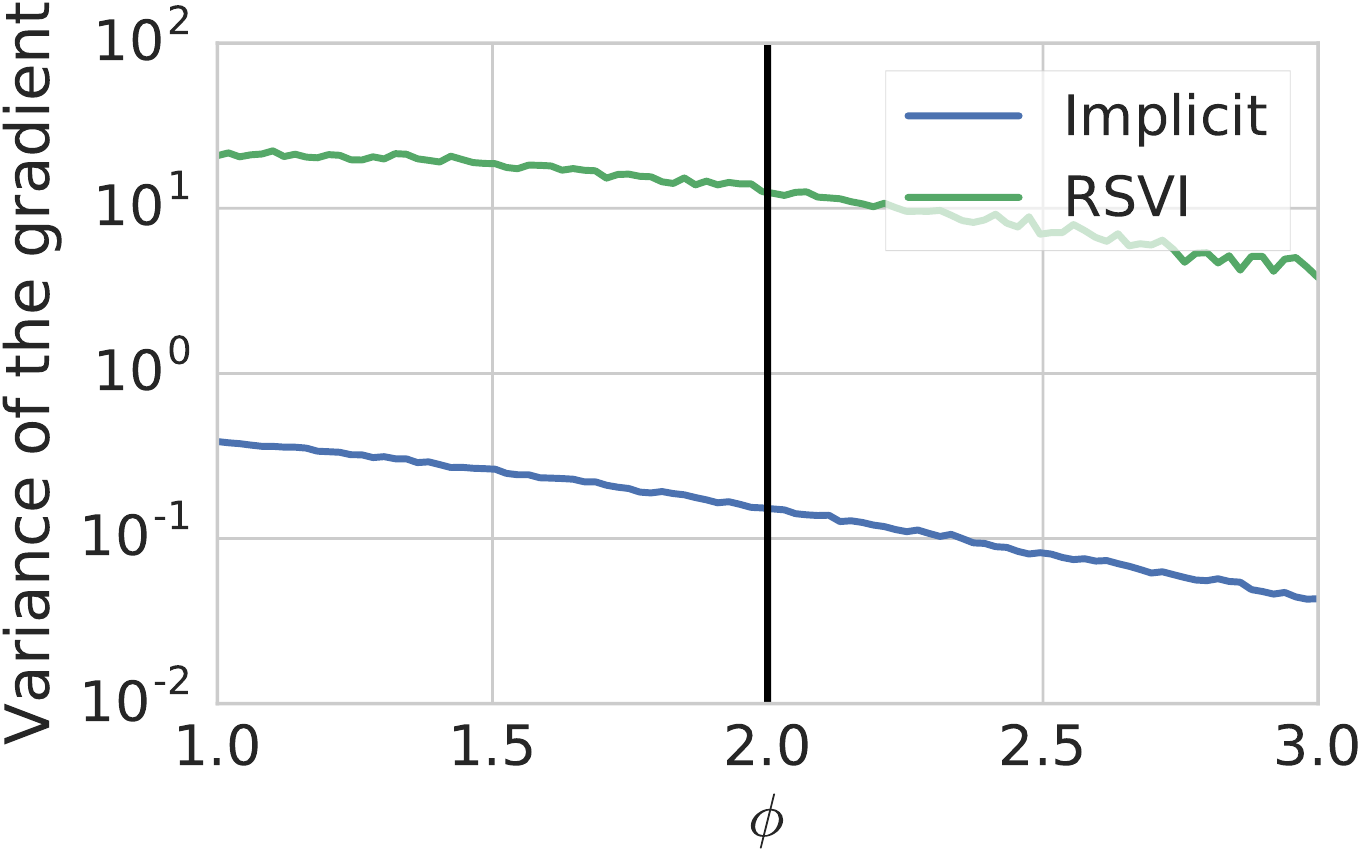}
        \caption{Von Mises distribution}
    \end{subfigure}
    \begin{subfigure}[b]{0.35\linewidth}
    \centering
    
    \resizebox{\linewidth}{!}{
    \begin{tabular}{llcc}\toprule
         \multicolumn{2}{l}{Method} & Dirichlet  & Von Mises \\ \midrule
         Implicit & & \textbf{\num{5.8e-08}} & \textbf{\num{2.0e-07}} \\
         \multirow{4}{*}{RSVI} & $B=0$ & \num{1.4e-07} & \multirow{4}{*}{\num{3.0e-07}} \\
         & $B=1$ & \num{1.6e-07} & \\
         & $B=5$ & \num{1.8e-07} & \\
         & $B=10$ & \num{2.0e-07} & \\
     \bottomrule
    \end{tabular}
    }
    \caption{Computation time (in seconds per sample of Gamma/von Mises)}
    \end{subfigure}
    \caption{Variance of the gradient and computation time for the cross-entropy optimization problem. The vertical line denotes the optimal value for the parameter. Implicit gradient is faster and has lower variance than RSVI~\cite{naesseth2017reparameterization}.}
    \label{fig:cross-entropy-variance}
\end{figure}

The results presented on Fig.~\ref{fig:cross-entropy-variance} show that the implicit gradient is faster and has lower variance than RSVI.
For the Dirichlet distribution, increasing the shape augmentation parameter $B$ allows RSVI to asymptotically approach the variance of the implicit gradient.
However, this comes at an additional computational cost and requires tuning this parameter.
Furthermore, such a parameter is not available for other distributions, including von Mises.

\subsection{Latent Dirichlet Allocation}
\label{sec:lda}

LDA~\cite{blei2003latent} is a popular topic model that represents each document as a bag-of-words and finds a set of topics so that each document is well-described by a few topics.
It has been extended in various ways, \eg~\cite{blei2005correlated,blei2006dynamic}, and often serves as a testbed for approximate inference methods~\cite{teh2007collapsed,hoffman2010online,hoffman2013stochastic}.
LDA is a latent variable model with a likelihood $p_{\Phi} (\vec{w} | \vec{z}) = \prod_{i=1}^K \operatorname{Categorical}(w_i | \Phi \vec{z})$, and the prior $p_{\vec{\alpha}} (\vec{z}) = \operatorname{Dirichlet}(\vec{z} | \vec{\alpha})$, where $\vec{w}$ is the observed document represented as a vector of word counts, $\vec{z}$ is a distribution of topics, $\Phi \in \mathbb{R}^{\text{\#words} \times \text{\#topics}}$ is a matrix that specifies the categorical distribution of words in each topic, and $\vec{\alpha}$ parameterizes the prior distribution over the topics.
We perform amortized variational inference by using a neural network to parameterize the Dirichlet variational posterior over the topics $\vec{z}$ as a function of the observation.

We use the 20 Newsgroups (11,200 documents, 2,000-word vocabulary) and RCV1~\cite{lewis2004rcv1} (800,000 documents, 10,000-word vocabulary) datasets with the same preprocessing as in~\cite{srivastava2017autoencoding}.
We report the test perplexity of the models, $\exp\left(-\frac{1}{N} \sum_{n=1}^N \frac{1}{L_n} \log p(\vec{w}_n)\right)$, where $L_n$ is the number of words in the document and the marginal log-likelihood is approximated with a single-sample estimate of the evidence lower bound.
Following~\cite{wallach2009rethinking}, we optimize the prior parameters $\vec{\alpha}$ during training.

We compare amortized variational inference in LDA using implicit reparameterization to several alternatives:
(i) training the LDA model with the RSVI gradients;
(ii) stochastic variational inference (SVI)~\cite{hoffman2013stochastic} training method for LDA;
(iii) the  method of~\citet{srivastava2017autoencoding}, which we refer to as LN-LDA, that uses a Logistic Normal approximation in place of the Dirichlet prior and performs amortized variational inference using a Logistic Normal variational posterior.

\begin{figure}
    \begin{subfigure}[b]{0.24\linewidth}
\tiny
$\alpha=1.15$ write article get think go \\
$\alpha=1.07$ write get think like article \\
$\alpha=1.07$ write article get think like \\
$\alpha=1.07$ write article get like know \\
$\alpha=1.06$ write article think get like \\
$\alpha=1.04$ write article get know think \\
$\alpha=1.04$ write article get know like \\
$\alpha=1.02$ write article think get like \\
        \caption{LN-LDA topics}
        \label{fig:lda-topics-20news-normal}
    \end{subfigure}
    \begin{subfigure}[b]{0.26\linewidth}
\tiny
$\alpha=0.47$ write article get like one \\
$\alpha=0.31$ write one people say think \\
$\alpha=0.25$ please thanks post send know \\
$\alpha=0.11$ use drive card problem system \\
$\alpha=0.10$ go say people know get \\
$\alpha=0.08$ use file key program system \\
$\alpha=0.08$ gun government law state use \\
$\alpha=0.08$ god christian jesus say people \\
        \caption{LDA topics (implicit)}
        \label{fig:lda-topics-20news-dirichlet}
    \end{subfigure}
    \begin{subfigure}[b]{0.24\linewidth}
        \includegraphics[width=\linewidth]{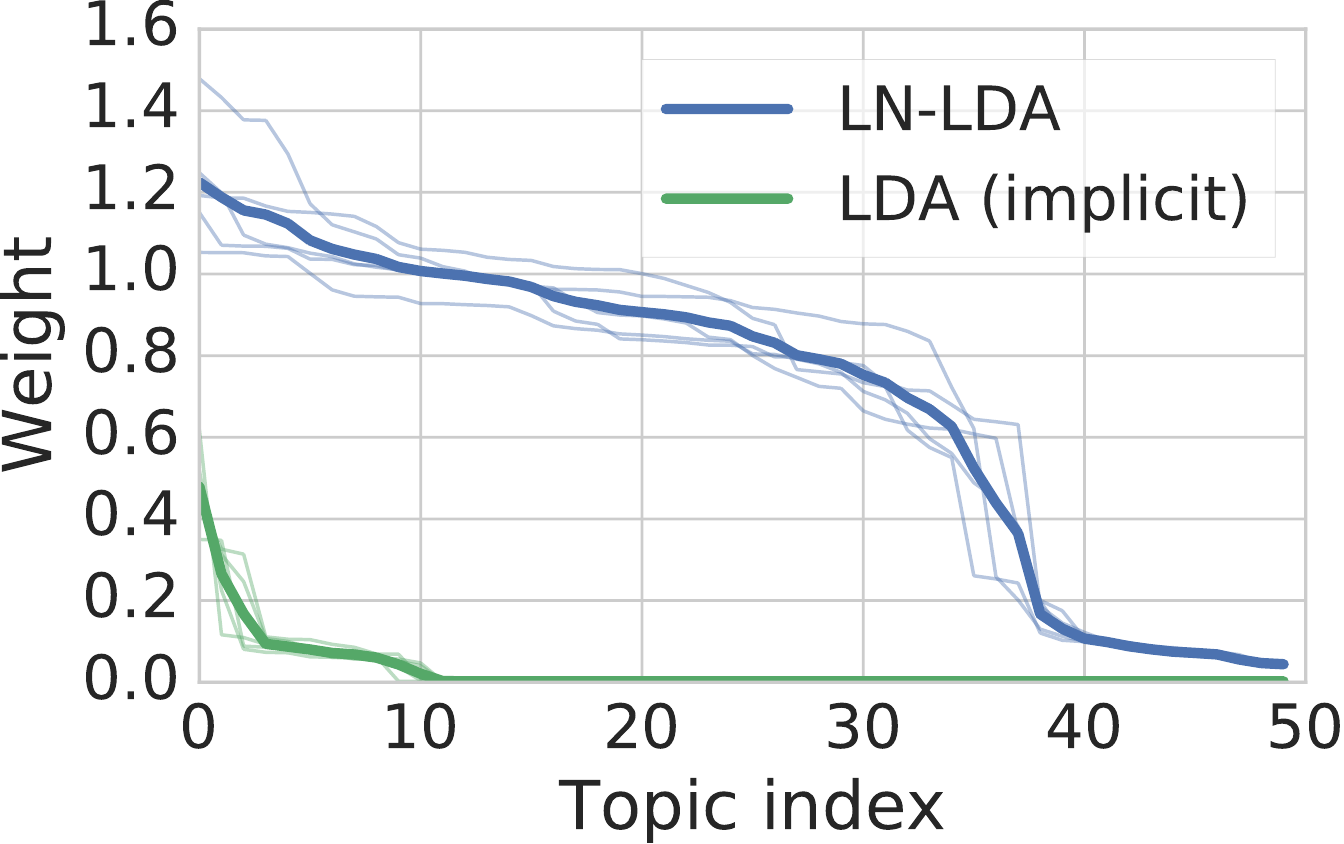}
        \caption{20 Newsgroups weights}
        \label{fig:lda-topic-weights-20news}
    \end{subfigure}
    \begin{subfigure}[b]{0.24\linewidth}
        \includegraphics[width=\linewidth]{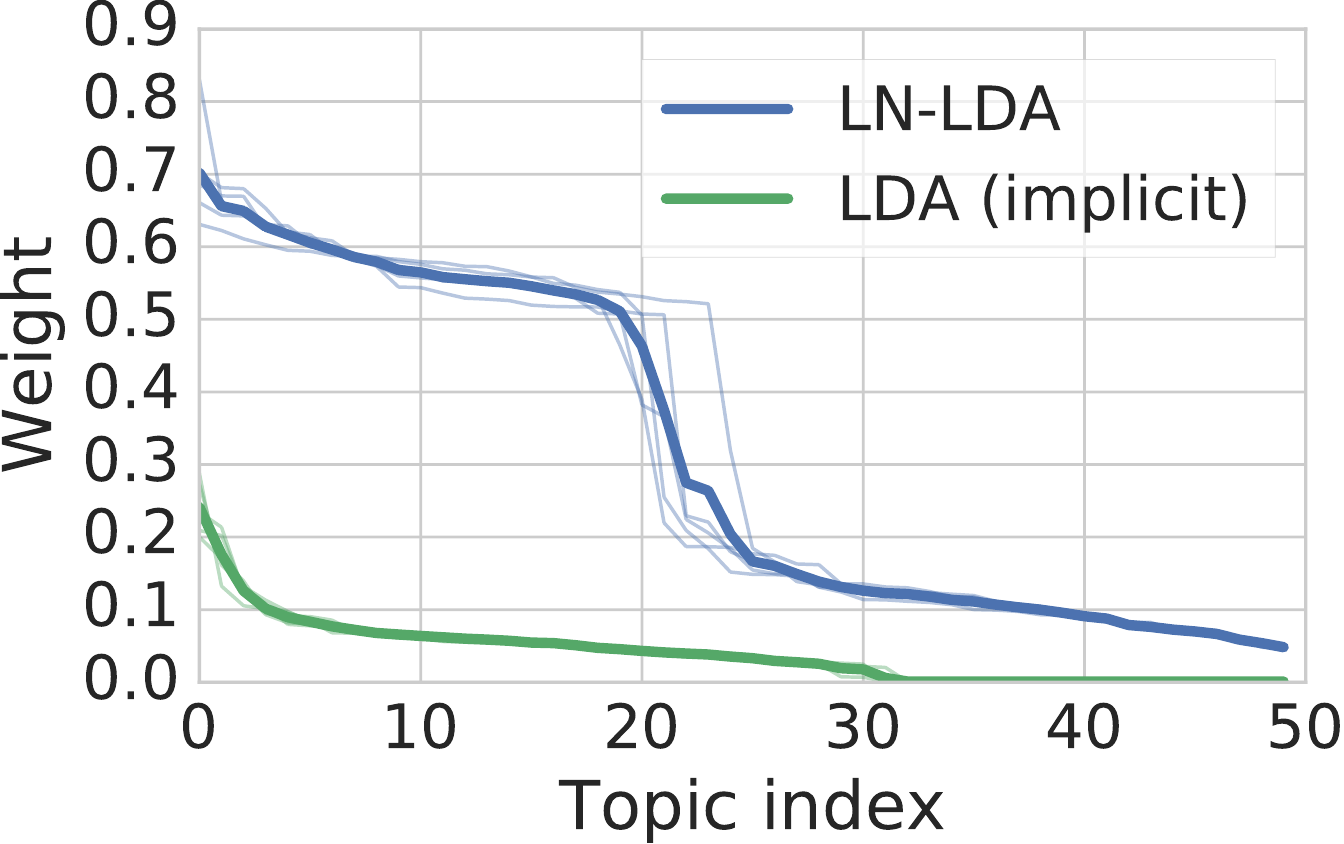}
        \caption{RCV1 weights}
        \label{fig:lda-topic-weights-rcv1}
    \end{subfigure}
    \caption{Left: topics with the highest weight for the 20 Newsgroups dataset; Right: prior topic weights $\vec{\alpha}$. LDA learns sparse prior weights, while LN-LDA does not.}
    \label{fig:lda-topics}
\end{figure}

\begin{figure}
    \centering
    \begin{subfigure}[b]{0.32\linewidth}
        \includegraphics[width=\linewidth]{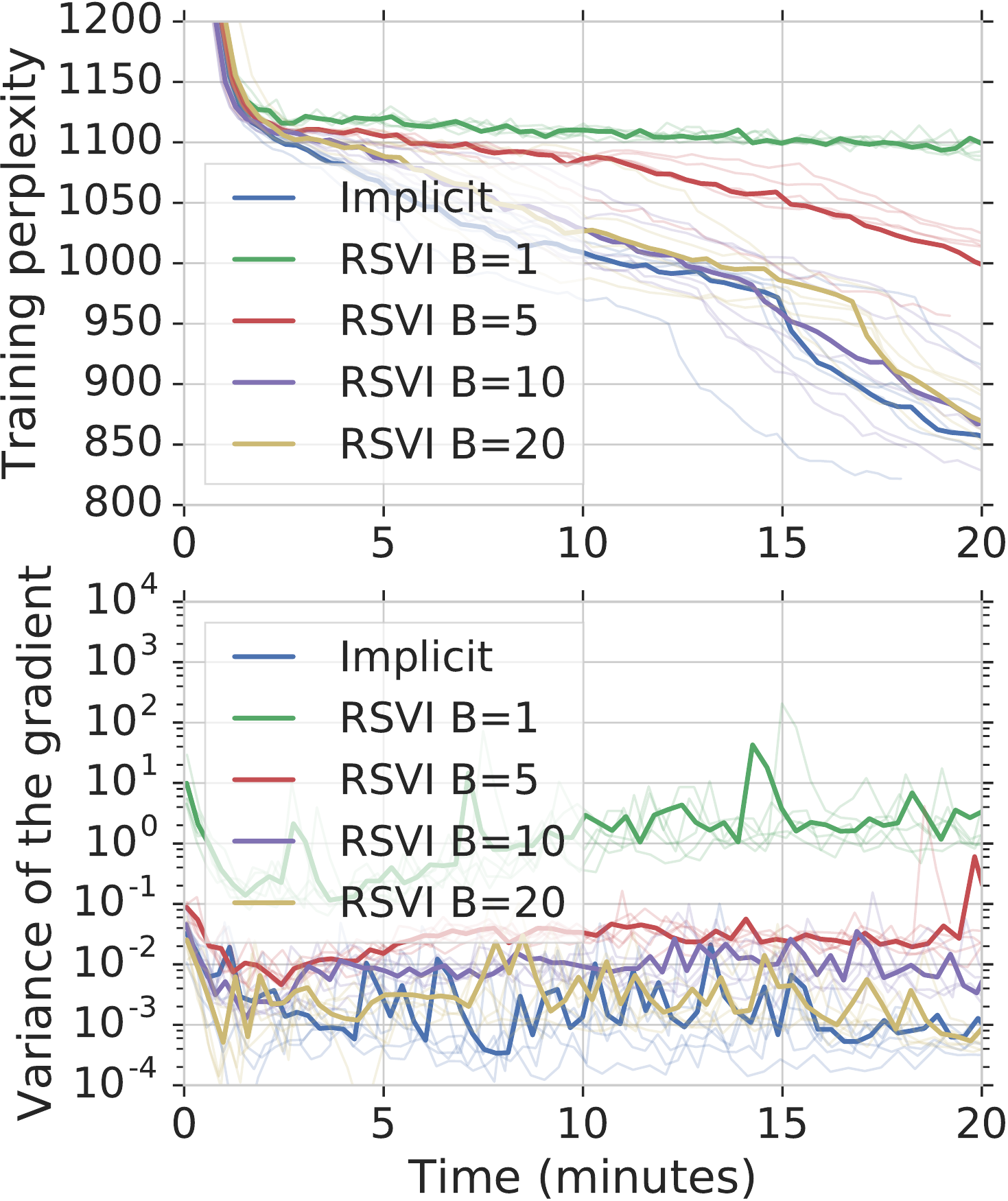}
        \caption{LDA on 20 Newsgroups}
    \end{subfigure}
    \begin{subfigure}[b]{0.32\linewidth}
        \includegraphics[width=\linewidth]{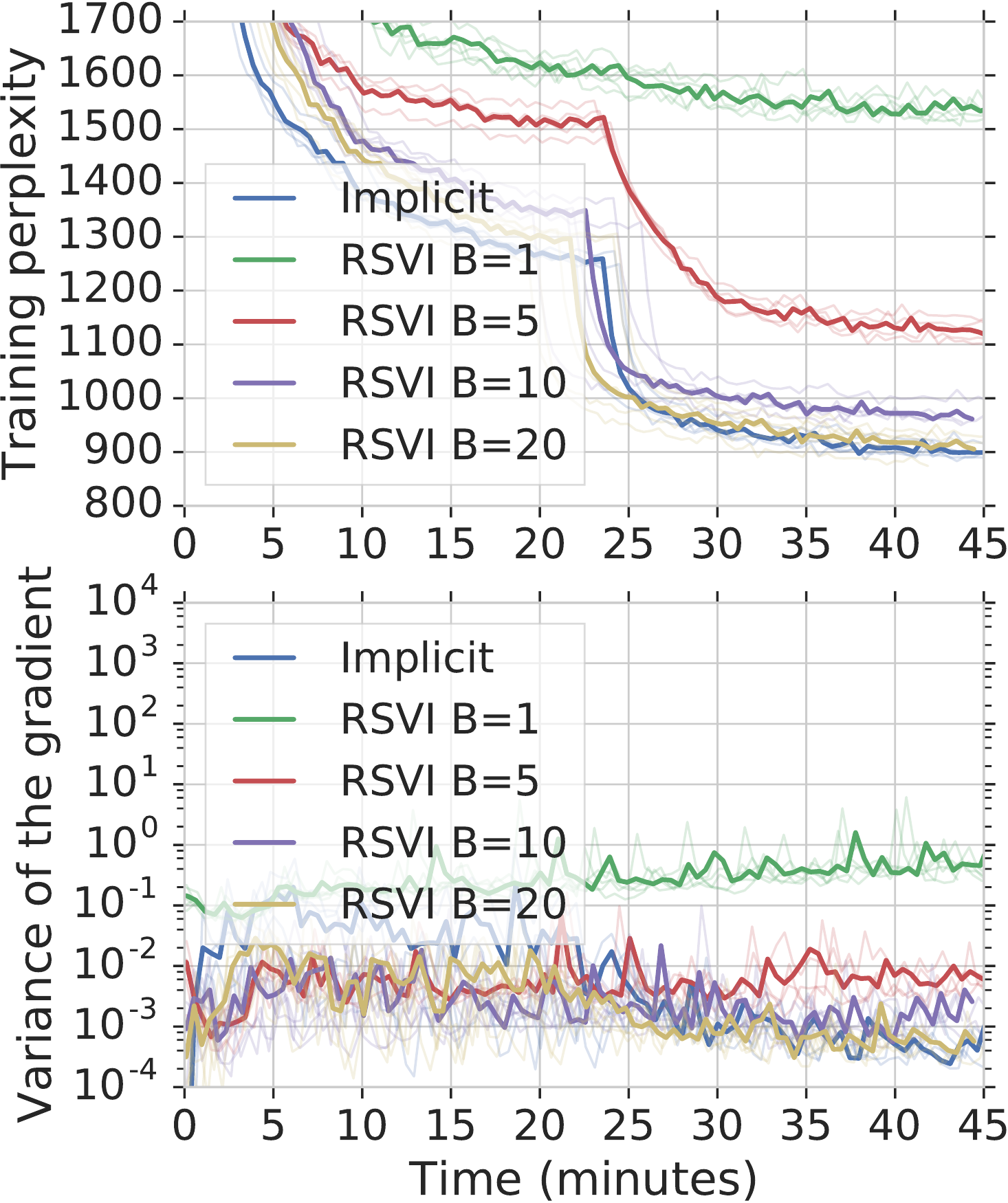}
        \caption{LDA on RCV1}
    \end{subfigure}
    \begin{subfigure}[b]{0.32\linewidth}
        \includegraphics[width=\linewidth]{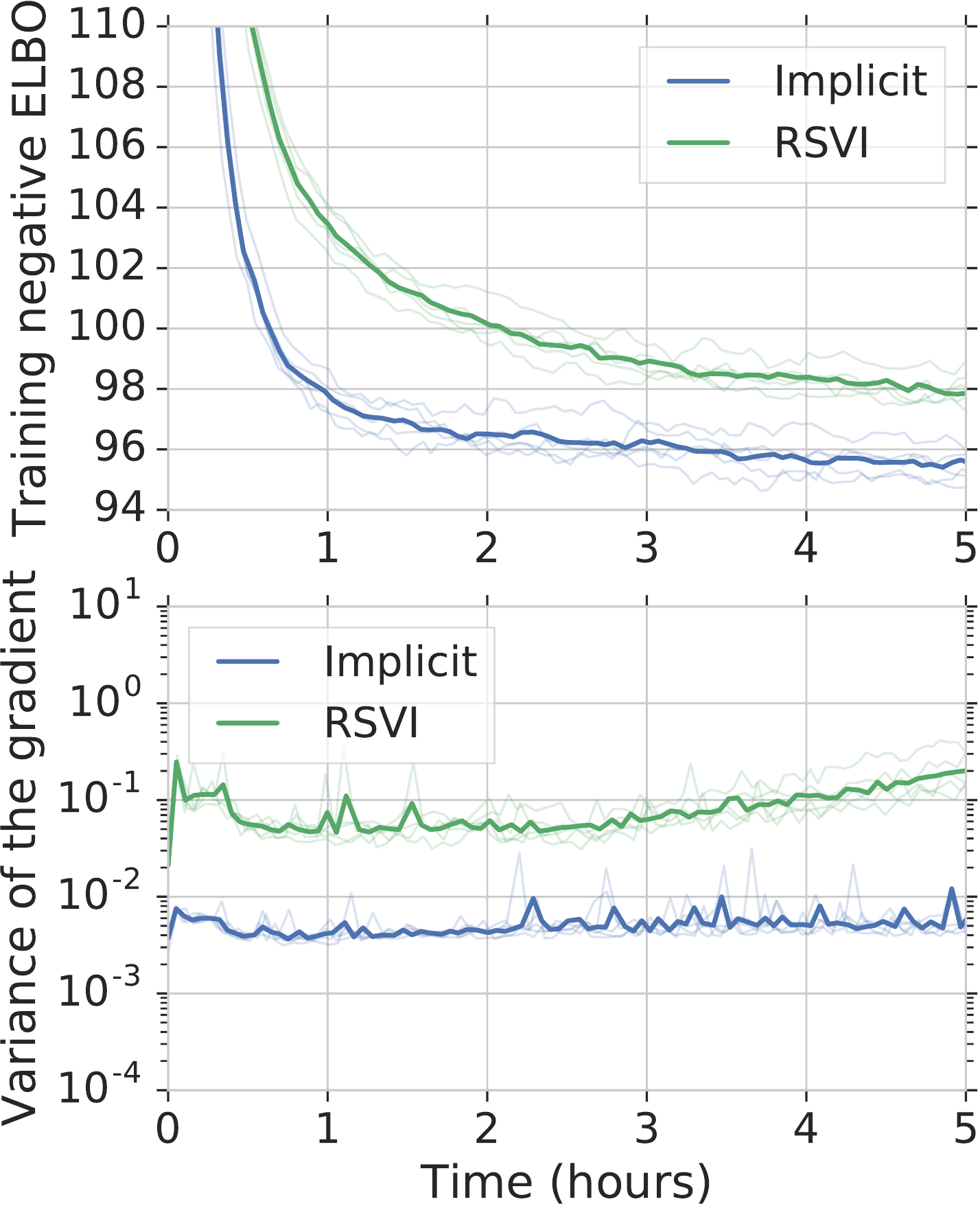}
        \caption{VAE with von Mises posterior}
        \label{fig:vonmises-vae-grad-variance}
    \end{subfigure}
    \caption{The training objective (top) and the variance of the gradient (bottom) during training. The sharp drop in perplexity on RCV1 dataset occurs at the end of the $\vec{\alpha}$ burn-in period.}
    \label{fig:grad-variance}
\end{figure}

The results in Table~\ref{fig:topic-modeling-ppl} and Fig.~\ref{fig:grad-variance}(a-b) show that RSVI matches the implicit gradient results only at $B=20$, as opposed to $B=10$ for the previous problem.
Lower gradient variance leads to faster training objective convergence.
Interestingly, amortized inference can achieve better perplexity than SVI.
Finally, we see that LDA trained with implicit gradients performs as well or better than LN-LDA.
The learned topics and the prior weights shown on Fig.~\ref{fig:lda-topics} demonstrate that LDA automatically determines the number of topics in the corpus by setting some of the prior weights to 0; this does not occur in LN-LDA model.
Additionally, LN-LDA is prone to representing the same topic several times, perhaps due to a non-sparse variational posterior distribution.

The obtained results suggest that the advantage of implicit gradients compared to RSVI increases with the complexity of the problem.
When the original distributions are replaced by surrogates, some desirable properties of the solution, such as sparsity, might be lost.

\begin{table}
    \centering
    \caption{Test perplexity (lower is better) for the topic modeling task. Mean $\pm$ standard deviation over 5 runs. LN-LDA uses Logistic Normal distributions instead of Dirichlet.}
    \resizebox{0.65\linewidth}{!}{
    \begin{tabular}{llcc}\toprule
         Model & Training method & 20 Newsgroups  & RCV1 \\ \midrule
         \multirow{6}{*}{LDA~\cite{blei2003latent}} & Implicit reparameterization & $876 \pm 7$ & $896 \pm 6$ \\
         &  RSVI $B=1$ & $1066 \pm 7$ & $1505 \pm 33$ \\
         & RSVI $B=5$ & $968 \pm 18$ & $1075 \pm 15$ \\
         & RSVI $B=10$ & $887 \pm 10$ & $953 \pm 16$ \\
         & RSVI $B=20$ & $865 \pm 11$ & $907 \pm 13$ \\
         & SVI & $964 \pm 4$ & $1330 \pm 4$ \\ \midrule
         LN-LDA~\cite{srivastava2017autoencoding} & Explicit reparameterization & $875 \pm 6$  & $951 \pm 10$ \\
     \bottomrule
    \end{tabular}
    }
    \label{fig:topic-modeling-ppl}
\end{table}

\subsection{Variational Autoencoders}

VAE~\cite{kingma2014auto,rezende2014stochastic} is a generative latent variable model trained using amortized variational inference.
Both the variational posterior and the generative distributions (also known as the encoder and decoder) are parameterized using neural networks. VAEs typically use the standard Normal distribution as the prior and a factorized Normal as the variational posterior. The form of the likelihood depends on the data, with factorized Bernoulli or Normal distributions being popular choices for images. 
In this section, we experiment with using distributions other than Normal for the prior and the variational posterior.
The use of alternative distributions allows incorporating different prior assumptions about the latent factors of the data, such as bounded support or periodicity.

We use fully factorized priors and variational posteriors.
For the variational posterior we explore Gamma, Beta, and von Mises distributions.
For Gamma, we use a sparse $\operatorname{Gamma}(0.3, 0.3)$ prior and a bell-shaped prior $\operatorname{Gamma}(10, 10)$.
For Beta and von Mises, instead of a sparse prior we choose a uniform prior over the corresponding domain.

\begin{figure}
    \centering
    \captionsetup{justification=centering}
    \begin{subfigure}[b]{0.27\linewidth}
        \centering
        \includegraphics[width=\linewidth]{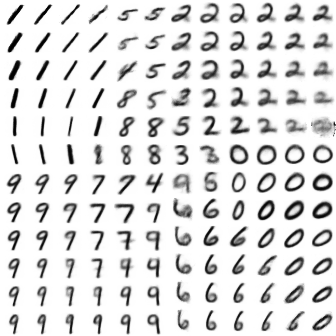}
        \caption{Normal posterior and prior, \\$[-3, 3] \times [-3, 3]$}
    \end{subfigure}\hfill
    \begin{subfigure}[b]{0.27\linewidth}
        \centering
        \includegraphics[width=\linewidth]{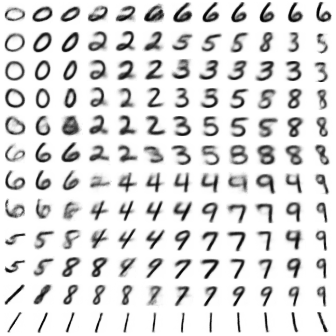}
        \caption{Beta, uniform prior, \\$[0, 1] \times [0, 1]$}
    \end{subfigure}\hfill
    \begin{subfigure}[b]{0.27\linewidth}
        \centering
        \includegraphics[width=\linewidth]{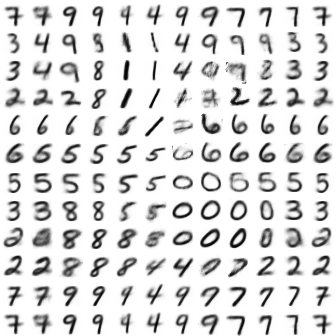}
        \caption{Von Mises, uniform prior, \\$[-\pi, \pi] \times [-\pi, \pi]$}
    \end{subfigure}
    \captionsetup{justification=justified}
    \caption{2D latent spaces learned by a VAE on the MNIST dataset. Normal distribution exhibits a strong pull to the center, while Beta and Von Mises latents are tiling the entire available space.}
    \label{fig:vae-mnist-latents}
\end{figure}

We train the models on the dynamically binarized MNIST dataset~\cite{burda2016importance} using the fully-connected encoder and decoder architectures from~\cite{davidson2018hyperspherical}, so our results are comparable.
The results in Table~\ref{fig:vae-mnist} show that a uniform prior and cyclic latent space of von Mises is advantageous for low-dimensional latent spaces, consistent with the findings of~\cite{davidson2018hyperspherical}.
For a uniform prior, the factorized von Mises distribution outperforms the multivariate von Mises-Fisher distribution in low dimensions, perhaps due to the more flexible concentration parameterization (von Mises-Fisher uses shared concentration across dimensions).
The results obtained with bell-shaped priors are similar to the Normal prior/posterior pair, as expected.
The latent spaces learned by models with 2 latents shown on Fig.~\ref{fig:vae-mnist-latents} demonstrate the differences in topology.

We provide a detailed comparison between implicit gradients and RSVI in Table~\ref{fig:vae-mnist-extended} of the supplementary material.
For Gamma and Beta distributions, RSVI with $B=20$ performs similarly to implicit gradients.
However, for the von Mises distribution implicit gradients usually perform better than RSVI. For example, for a uniform prior and $D = 40$, implicit gradients yield a $1.3$ nat advantage in the test log-likelihood due to lower gradient variance (Fig.~\ref{fig:vonmises-vae-grad-variance}).

\begin{table}
    \centering
    \caption{Test negative log-likelihood (lower is better) for VAE on MNIST. Mean $\pm$ standard deviation over 5 runs. The von Mises-Fisher results are from~\cite{davidson2018hyperspherical}.}
    \resizebox{\linewidth}{!}{
    \begin{tabular}{llccccc}\toprule
        Prior & Variational posterior & $D=2$ & $D=5$ & $D=10$ & $D=20$ & $D=40$ \\ \midrule
        $\mathcal{N}(0, 1)$ & $\mathcal{N}(\mu, \sigma^2)$ & $131.1 \pm 0.6$ & $107.9 \pm 0.4$ & $92.5 \pm 0.2$ & $88.1 \pm 0.2$ & $88.1 \pm 0.0$ \\ \midrule
        $\operatorname{Gamma}(0.3, 0.3)$ & $\operatorname{Gamma}(\alpha, \beta)$ & $132.4 \pm 0.3$ & $108.0 \pm 0.3$ & $94.0 \pm 0.3$ & $90.3 \pm 0.2$ & $90.6 \pm 0.2$ \\
        $\operatorname{Gamma}(10, 10)$ & $\operatorname{Gamma}(\alpha, \beta)$ & $135.0 \pm 0.2$ & $107.0 \pm 0.2$ & $92.3 \pm 0.2$ & $88.3 \pm 0.2$ & $88.3 \pm 0.1$ \\
        $\operatorname{Uniform}(0, 1)$ & $\operatorname{Beta}(\alpha, \beta)$ & $128.3 \pm 0.2$ & $107.4 \pm 0.2$ & $94.1 \pm 0.1$ & $88.9 \pm 0.1$ & $88.6 \pm 0.1$ \\
        $\operatorname{Beta}(10, 10)$ & $\operatorname{Beta}(\alpha, \beta)$ & $131.1 \pm 0.4$ & $\mathbf{106.7} \pm 0.1$ & $\mathbf{92.1} \pm 0.2$ & $\mathbf{87.8} \pm 0.1$ & $\mathbf{87.7} \pm 0.1$ \\
        $\operatorname{Uniform}(-\pi, \pi)$ & $\operatorname{vonMises}(\mu, \kappa)$ & $\mathbf{127.6} \pm 0.4$ & $107.5 \pm 0.4$ & $94.4 \pm 0.5$ & $90.9 \pm 0.1$ & $91.5 \pm 0.4$ \\
        $\operatorname{vonMises}(0, 10)$ & $\operatorname{vonMises}(\mu, \kappa)$ & $130.7 \pm 0.8$ & $107.5 \pm 0.5$ & $92.3 \pm 0.2$ & $\mathbf{87.8} \pm 0.2$ & $87.9 \pm 0.3$ \\ \midrule
        $\operatorname{Uniform}(S^D)$ & $\operatorname{vonMisesFisher}(\vec{\mu}, \kappa)$ & $132.5 \pm 0.7$ & $108.4 \pm 0.1$ & $93.2 \pm 0.1$ & $89.0 \pm 0.3$ & $90.9 \pm 0.3$ \\
     \bottomrule
    \end{tabular}
    }
    \label{fig:vae-mnist}
\end{table}

\section{Conclusion}
Reparameterization gradients have become established as a central tool underlying many of the recent advances in machine learning. In this paper, we strengthened this tool by extending its applicability to distributions, such as truncated, Gamma, and von Mises, that are often encountered in probabilistic modelling. The proposed implicit reparameterization gradients offer a simple and practical approach to stochastic gradient estimation which has the properties we expect from such a new type of estimator: it is faster than the existing methods and simultaneously provides lower gradient variance. These new estimators allow us to move away from making model choices for reasons of computational convenience. Applying these estimators requires a numerically tractable CDF or some other standardization function. When one is not available, it should be possible to use an approximate standardization function to augment implicit reparameterization with a score function correction term, along the lines of generalized reparameterization. We intend to explore this direction in future work.

\subsubsection*{Acknowledgments}
We would like to thank Chris Maddison, Hyunjik Kim, J\"org Bornschein, Alex Graves, Hussein Fawzi, Chris Burgess, Matt Hoffman, and Charles Sutton for helpful discussions.
We also thank Akash Srivastava for providing the preprocessed document datasets.

\printbibliography

\newpage
\appendix

\begin{refsection}
\newpage
\appendix

\begin{center}
    \textbf{\Large Implicit Reparameterization Gradients \\ Supplementary materials}
\end{center}

\section{Testing of implicit gradient implementation}
A simple test to verify the correctness of an implicit gradient implementation is to choose an appropriate function $f(\vec{z})$ and check that the Monte-Carlo averaging approximates the correct quantity:
\begin{equation}
    \nabla_{\vec{\phi}} \E_{q_{\vec{\phi}} (\vec{z})} \left[f(\vec{z})\right] \approx \frac{1}{S} \sum_{s=1}^S \nabla_{\vec{z}}  f(\vec{z_s}) \nabla_{\vec{\phi}} \vec{z}_s,\ \vec{z}_s \sim q_{\vec{\phi}}(\vec{z}).
\end{equation}
For $\operatorname{Gamma}(\alpha, 1)$, we can choose $f(z) = z$.
Then, $\frac{\partial}{\partial \alpha} \E_{q_{\alpha}(z)} z = 1$, so the average stochastic gradient should be equal to 1.
For $\operatorname{vonMises}(0, \kappa)$, an appropriate choice is $f(z) = \cos z$.
Then, $\frac{\partial }{\partial \kappa} \E_{q_{\kappa}(z)} \cos z = \frac{\partial}{\partial \kappa} \frac{I_1(\kappa)}{I_0(\kappa)} = 1 - \frac{I_1(\kappa)}{\kappa I_0(\kappa)} - \left( \frac{I_1(\kappa)}{I_0(\kappa)} \right)^2$.

\section{Implementation details for the reparameterization gradient}
\label{sec:reparameterization-gradient-implementation}

We implement Eqn.~\eqref{eqn:z-grad-univariate} using an equivalent but more numerically stable expression:
\begin{equation}
    \nabla_{\vec{\phi}} z = - \frac{\nabla_{\vec{\phi}} F(z | \vec{\phi})}{q_{\vec{\phi}} (z)} = -\exp( - \log q_{\vec{\phi}} (z)) \nabla_{\vec{\phi}} F(z | \vec{\phi}).
\end{equation}

\textbf{Gamma distribution.}
We perform forward-mode differentiation of the efficient computation method~\cite{bhattacharjee1970incgamma}.
We use the implementation available in Eigen~\cite{eigenweb}, which is based on the Cephes~\cite{moshier2000cephes} library. A more advanced version of this method is available in SciPy~\cite{jones2001scipy}.
For $z \geq 1$ and $z > \alpha$ this method uses the continued fraction expansion:
\begin{equation}
    \gamma(z, \alpha) = 1 - \frac{\exp(-z) z^\alpha}{\Gamma(\alpha)} \cfrac{1}{z + \cfrac{1 - \alpha}{1+ \cfrac{1}{z + \cfrac{2 - \alpha}{1 + \frac{2}{z+ \dots}}}}}
\end{equation}
The expansion can be evaluated in the ``direct order'' using the Wallis algorithm~\cite{press2007numerical}.
For other values of the arguments, a series expansion is used:
\begin{equation}
    \gamma(z, \alpha) = \frac{\exp(-z) z^\alpha}{\Gamma(\alpha + 1)} \left( 1 + \sum_{k=1}^\infty \frac{z^k}{(\alpha+1) (\alpha+2) \dots (\alpha+r)} \right)
\end{equation}
In both cases, all the operations are differentiable with respect to $\alpha$, so forward-mode differentiation can be applied.
We stop the computation as soon as the value of the derivative (not the CDF) starts changing by less than some small value.
The maximum number of iterations is set to 200 for \texttt{float32} precision and 500 for \texttt{float64}.
Additionally, we multiply the resulting derivative by $\exp(- \log \operatorname{Gamma}(z|\alpha, 1))$ in the same code.
This improves the speed by about 30\% and the accuracy by an order of magnitude, because some of the terms, including a Gamma function, cancel out.

\textbf{Von Mises distribution.}
The CDF of standardized von Mises distribution is given by the series
\begin{equation}
    F(z | 0, \kappa) = \int_{-\pi}^z \operatorname{vonMises}(t | 0, \kappa) dt = \frac{z}{2\pi} + \frac{1}{\pi} \sum_{j=1}^\infty \frac{I_j(\kappa)}{I_0(\kappa)} \frac{\sin(j \cdot z)}{j}.
    \label{eqn:vm-series}
\end{equation}
For smaller concentration parameters, $\kappa < 50$, the numerical method~\cite{hill1977algorithm} first chooses the truncation point $K$ for the series, and then computes the first $K$ terms using an efficient backwards recursion.
For larger $\kappa$, it computes the CDF of a Normal approximation for the von Mises.
We use the implementation available in the SciPy library~\cite{jones2001scipy}.
Again, forward-mode automatic differentiation can be used since all the operations with respect to $\kappa$ are differentiable.

\section{Accuracy and speed of the reparameterization gradient estimators}
\label{sec:accuracy-speed-appendix}

We start by describing how we computed the ground-truth value of the CDF derivatives and then
provide the implementation details of the comparison.

\textbf{Gamma distribution.}
The derivative of the CDF of Gamma distribution can be obtained in terms of the hypergeometric function ${}_2 F_2$~\cite{wolframgammader}:
\begin{equation}
    \frac{\partial \gamma(z, \alpha)}{\partial \alpha} = \gamma(z, \alpha) (\log z - \psi(\alpha)) + {}_2 F_2 (\alpha, \alpha; \alpha + 1, \alpha + 1; -z) \frac{z^{\alpha}}{\alpha \Gamma(\alpha + 1)},
\end{equation}
where $\psi(\alpha) = (\log \Gamma(\alpha))'$ is the digamma function and ${}_2 F_2 (\alpha, \alpha; \alpha + 1, \alpha + 1; -z) = \sum_{k=0}^\infty \frac{\alpha^2}{(\alpha + k)^2} \frac{(-z)^k}{k!}$.
The function ${}_2 F_2$ is implemented in \verb|mpmath| package~\cite{mpmath}, allowing to evaluate this expression to arbitrary precision for comparison purposes.
We compute it with the default settings that result in \texttt{float64} precision.

\textbf{Von Mises distribution.}
Differentiating the series~\eqref{eqn:vm-series} with respect to $\kappa$ using the identity $\frac{\partial}{\partial \kappa} I_j(\kappa) = \frac{j}{\kappa} I_j(\kappa) + I_{j+1}(\kappa)$ yields
\begin{equation}
    \frac{\partial F(z | 0, \kappa)}{\partial \kappa} =  \frac{1}{\pi} \sum_{j=1}^\infty \left( \frac{\frac{j}{\kappa} I_j(\kappa) + I_{j+1} (\kappa)}{I_0(\kappa)} - \frac{I_j(\kappa) I_1(\kappa)}{(I_0(\kappa))^2} \right) \frac{\sin(j \cdot z)}{j}.
\end{equation}
We compute this expression using the SciPy implementation of the Bessel functions by truncating the series at the $100^{th}$ term.

\textbf{Details of the comparison.}
We first choose a grid of the parameters.
For the Gamma distribution, we consider $\alpha \in \{\num{1e-2}, \num{1e-1}, \num{1}, \num{1e1}, \num{1e2}, \num{1e3}\}$, and for von Mises we consider $\kappa \in \{ \num{1e-2}, \num{1e-1}, \num{1}, \num{1e1} \}$.
Then, we sample $1000$ random variables from the distribution for each value of the parameter.
We report the relative step sizes $\delta > 0$ determined by grid search that we use for the finite difference approximation in Table~\ref{tbl:finite-difference-stepsize}.
The timings are measured on a single core of an Intel Xeon CPU.

\begin{table}
    \centering
    \caption{The relative step size $\delta$ used for finite difference approximation of the CDF derivative.}
    \begin{tabular}{lcc}\toprule
        & \texttt{float32} & \texttt{float64} \\ \midrule
        Gamma & $10^{-3}$ & $10^{-5}$ \\
        Von Mises & $10^{-1}$ & $10^{-4}$ \\
        \bottomrule
    \end{tabular}
    \label{tbl:finite-difference-stepsize}
\end{table}

\section{Experimental details}
\label{sec:experimental-details}

\textbf{RSVI details.}
We use the proposal distributions suggested in~\cite{naesseth2017reparameterization}.
For $\operatorname{Gamma}(\alpha, 1)$, we employ the proposal distribution from~\citet{marsaglia2000simple} which is explicitly reparameterizable using the standard Normal.
For $\operatorname{vonMises}(0, \kappa)$, we use the wrapped Cauchy proposal distribution~\cite{best1979efficient} that is explicitly reparameterizable using the Uniform distribution.

\textbf{Gradient of the cross-entropy.}
The gradient variance is computed as
\begin{equation}
    \mathbb{E}_{q_{\phi}(\vec{z})} \left( \frac{\partial}{\partial \phi} [-\log p(\vec{z})] - c \right)^2,\
\end{equation}
where the expectation is estimated using $1000$ samples and $c = \frac{\partial}{\partial \phi} \mathbb{E}_{q_{\phi}(\vec{z})} [-\log p(\vec{z})]$ is the analytical gradient of the cross-entropy.
The timings are measured on a single core of an Intel Xeon CPU.

\textbf{Variance of the gradient during training.}
For LDA and VAE models, we estimate the variance of the gradient by reusing the exponential moving averages of the first and second moments computed by the Adam optimizer~\cite{kingma2015adam}.
Specifically, denoting by $\vec{m}$ and $\vec{v}$ the estimates of the first and second moments of the gradient respectively, the variance estimate is $(\vec{v} - \vec{m}^2)$.
We average this estimate over all the parameters.
Note that we compute the variance of the gradient with respect to the model parameters: weights, biases and prior parameters.

\textbf{Parameters of distributions.}
The positive-valued parameters (scale of Normal; all the parameters of Gamma, Beta and Dirichlet; concentration parameter of von Mises) are computed as a softplus of an unconstrained value.
For all of these parameters except for the Normal scale, we additionally perform clipping to the $[10^{-3}, 10^{3}]$ range.
This makes the analytical KL-divergence numerically stable in \texttt{float32} precision.
A useful sanity check of numerical stability is that the KL-divergence is always non-negative.

The samples from von Mises distribution, and likewise the location parameter $\mu$, can be equivalently represented as an angle $z \in [-\pi, \pi)$, or as a point on a circle $(x, y)$.
We find that learning in the second case is much easier.
Thus, we compute the location parameter as $\mu = \texttt{atan2}(x, y)$, where $x$ and $y$ are unconstrained values, and transform the samples from the distribution: $z \to (\cos z, \sin z)$.

\textbf{Latent Dirichlet Allocation.}
The 20 Newsgroups models are trained for 500 epochs, while the RCV1 ones for 10 epochs.
We set the number of topics to 50.
The inference network is a ReLU multilayer perceptron with the same number of units per layer.
The optimization method is Adam~\cite{kingma2015adam} with $\beta_1 = 0.9$ and the batch size is $32$.
The model parameters are initialized using the Xavier initializer~\cite{glorot2010understanding}: the truncated Normal distribution with zero mean and the variance of $\frac{2}{\texttt{fan\_in} + \texttt{fan\_out}}$.
The prior parameters are fixed at the initial value for a number of epochs (burn-in period) and then trained jointly with other parameters.
For the LN-LDA model~\cite{srivastava2017autoencoding}, we tune the prior parameters of the underlying Dirichlet distribution for which the Laplace approximation is performed; we also checked that training the Normal prior parameters without any constraints does not improve the perplexity.
We find that the architectural modifications suggested in~\cite{srivastava2017autoencoding}, such as using dropout and batch normalization, do not lead to improved values of the perplexity, so we do not use them (they report the perplexity of $1059$ for 20 Newsgroups dataset, while we obtain $875$).

We find the key hyperparameters by Bayesian optimization of the validation set perplexity.
The validation set consists of $10\%$ random training documents for 20 Newsgroups and $1\%$ random training documents for RCV1.
A separate search is performed for each dataset (20 Newsgroups, RCV1) and model (LN-LDA, LDA (implicit)) combination, for a total of four runs.
For 20 Newsgroups the obtained hyperparameters are very similar for both models, so we use the same values.
The hyperparameter values are shown in Table~\ref{tbl:lda-hyperparameters}.

We use GenSim library~\cite{gensim2010} implementation of stochastic variational inference (SVI).
We train for the same number of epochs and with the same number of topics as above.
We perform a grid search for the key hyperparameters.
For 20 Newsgroups, we use \texttt{chunk\_size=1000} and \texttt{decay=0.5}, while for RCV1 we set \texttt{chunk\_size=2000} and \texttt{decay=0.5}.
In both cases, we set \texttt{alpha="auto"}, meaning that the prior hyperparameters $\vec{\alpha}$ are learned.
The remaining options were set to the default values.

We performed control experiments on the 20 Newsgroups dataset showing that (i) using a fixed prior distribution increases the perplexity by 60 points; (ii) computing the KL using sampling instead of an analytical expression increases the perplexity by 80 points.
The latter result highlights the importance of using variational posteriors that allow for analytical KL estimation when dealing with ``sparse'' distributions that have density asymptotes.

\begin{table}
    \centering
    \caption{Hyperparameters used for the LDA experiments.}
    \resizebox{0.7\linewidth}{!}{
    \begin{tabular}{lccc}\toprule
        Dataset & 20 Newsgroups & RCV1 & RCV1 \\
        Model & LN-LDA, LDA & LN-LDA & LDA \\ \midrule
        Learning rate & \num{3e-4} & \num{1e-3} & \num{1e-3} \\
        Layers in the inference network & 3 & 1 & 2 \\
        Units per layer & 300 & 200 & 250 \\
        Initial value of $\vec{\alpha}$ & 0.7 & 0.5 & 0.95 \\
        Burn-in epochs for $\vec{\alpha}$ & 350 & 7 & 5 \\
        \bottomrule
    \end{tabular}
    }
    \label{tbl:lda-hyperparameters}
\end{table}

\textbf{Variational autoencoder.}
We base our experimental setup on the one from~\citet{davidson2018hyperspherical}: a fully-connected ReLU network with two layers of 256 and 128 units as the encoder, a two-layer fully-connected ReLU network with 128 and 256 units as the decoder, minibatch size of 64, Adam optimizer~\cite{kingma2015adam}, and annealing the KL term from $0$ to $1$ over the first $10^5$ minibatches.
The only differences are (i) we do not perform early stopping and always train for 2 million minibatches; (ii) we train each model with the learning rates of $10^{-3}$ and $10^{-4}$ and choose the best-performing one.
The model parameters are initialized from the truncated Normal distribution with zero mean and the variance of $\frac{1}{\texttt{fan\_in}}$.
We estimate the log-likelihood using importance sampling with 500 samples.

We present the comparison between implicit gradients and RSVI gradients in Table~\ref{fig:vae-mnist-extended}.
We find that while they perform similarly for Gamma and Beta distributions, for the von Mises distribution implicit gradients obtain better results, since there is no analogue of the shape augmentation parameter $B$ for this distribution.

\begin{table}
    \centering
    \caption{Comparison of implicit reparameterization gradients and RSVI for the generative modeling task on MNIST dataset. Test negative log-likelihood (lower is better) mean $\pm$ standard deviation over 5 runs.}
    \resizebox{\linewidth}{!}{
    \begin{tabular}{lllccccc}\toprule
Prior & Variational posterior & Training method & $D=2$ & $D=5$ & $D=10$ & $D=20$ & $D=40$ \\ \midrule
$\mathcal{N}(0, 1)$ & $\mathcal{N}(\mu, \sigma^2)$ & Explicit & $131.1 \pm 0.6$ & $107.9 \pm 0.4$ & $92.5 \pm 0.2$ & $88.1 \pm 0.2$ & $88.1 \pm 0.0$ \\ \midrule
\multirow{2}{*}{$\operatorname{Gamma}(0.3, 0.3)$} & \multirow{2}{*}{$\operatorname{Gamma}(\alpha, \beta)$} & Implicit & $132.4 \pm 0.3$ & $108.0 \pm 0.3$ & $94.0 \pm 0.3$ & $90.3 \pm 0.2$ & $90.6 \pm 0.2$ \\
 &  & RSVI $B=20$ & $132.3 \pm 0.2$ & $108.5 \pm 0.3$ & $94.3 \pm 0.9$ & $90.1 \pm 0.1$ & $90.6 \pm 0.1$ \\ \midrule
\multirow{2}{*}{$\operatorname{Gamma}(10, 10)$} & \multirow{2}{*}{$\operatorname{Gamma}(\alpha, \beta)$} & Implicit & $135.0 \pm 0.2$ & $107.0 \pm 0.2$ & $92.3 \pm 0.2$ & $88.3 \pm 0.2$ & $88.3 \pm 0.1$ \\
 &  & RSVI $B=20$ & $131.6 \pm 0.3$ & $107.1 \pm 0.1$ & $92.2 \pm 0.1$ & $88.2 \pm 0.1$ & $88.2 \pm 0.1$ \\ \midrule
\multirow{2}{*}{$\operatorname{Uniform}(0, 1)$} & \multirow{2}{*}{$\operatorname{Beta}(\alpha, \beta)$} & Implicit & $128.3 \pm 0.2$ & $107.4 \pm 0.2$ & $94.1 \pm 0.1$ & $88.9 \pm 0.1$ & $88.6 \pm 0.1$ \\
 &  & RSVI $B=20$ & $128.9 \pm 0.8$ & $107.3 \pm 0.1$ & $94.3 \pm 0.1$ & $88.8 \pm 0.1$ & $88.5 \pm 0.1$ \\ \midrule
\multirow{2}{*}{$\operatorname{Beta}(10, 10)$} & \multirow{2}{*}{$\operatorname{Beta}(\alpha, \beta)$} & Implicit & $131.1 \pm 0.4$ & $106.7 \pm 0.1$ & $92.1 \pm 0.2$ & $87.8 \pm 0.1$ & $87.7 \pm 0.1$ \\
 &  & RSVI $B=20$ & $131.7 \pm 0.4$ & $106.9 \pm 0.1$ & $92.2 \pm 0.1$ & $87.7 \pm 0.1$ & $87.6 \pm 0.1$ \\ \midrule
\multirow{2}{*}{$\operatorname{Uniform}(-\pi, \pi)$} & \multirow{2}{*}{$\operatorname{vonMises}(\mu, \kappa)$} & Implicit & $127.6 \pm 0.4$ & $107.5 \pm 0.4$ & $94.4 \pm 0.5$ & $90.9 \pm 0.1$ & $91.5 \pm 0.4$ \\
 &  & RSVI & $129.1 \pm 0.4$ & $107.6 \pm 0.3$ & $96.0 \pm 0.5$ & $92.8 \pm 0.2$ & $92.8 \pm 0.2$ \\ \midrule
\multirow{2}{*}{$\operatorname{vonMises}(0, 10)$} & \multirow{2}{*}{$\operatorname{vonMises}(\mu, \kappa)$} & Implicit & $130.7 \pm 0.8$ & $107.5 \pm 0.5$ & $92.3 \pm 0.2$ & $87.8 \pm 0.2$ & $87.9 \pm 0.3$ \\
 &  & RSVI & $130.4 \pm 0.7$ & $107.8 \pm 0.5$ & $93.0 \pm 0.1$ & $88.7 \pm 0.2$ & $88.7 \pm 0.1$ \\
     \bottomrule
    \end{tabular}
    }
    \label{fig:vae-mnist-extended}
\end{table}
 \printbibliography
\end{refsection}

\end{document}